\documentclass[runningheads]{llncs}

\usepackage{accv}
\usepackage{accvabbrv}

\usepackage{graphicx}
\usepackage{booktabs}

\usepackage[accsupp]{axessibility}  

\usepackage[pagebackref,breaklinks,colorlinks,citecolor=accvblue]{hyperref}

\usepackage{orcidlink}

\usepackage{bbm}
\usepackage{algorithm}
\usepackage{algorithmic}

\usepackage{hyperref}       
\usepackage{url}            

\makeatletter
\def\blfootnote{\xdef\@thefnmark{}\@footnotetext}
\makeatother

\def\qi{\textcolor{black}}

\begin{document}

\title{Act Like a Radiologist: Radiology Report Generation across Anatomical Regions} 

\titlerunning{Act Like a Radiologist}

\author{Qi Chen\inst{1}$^*$\orcidlink{0000-0001-8732-8049} \and Yutong Xie\inst{1}$^*$\orcidlink{0000-0002-6644-1250} \and Biao Wu\inst{2} \and Xiaomin Chen\inst{3} \and James Ang\inst{4} \\ Minh-Son To\inst{1,4,5} \and Xiaojun Chang\inst{2} \and Qi Wu\inst{1}$^\dagger$}

\authorrunning{Qi Chen et al.}


\institute{Australian Institute for Machine Learning, University of Adelaide \and University of Technology Sydney, $^3$ South China University of Technology \\ $^4$ Royal Adelaide Hospital, $^5$ Flinders University}

\maketitle

\begin{abstract}
    Automating radiology report generation can ease the reporting workload for radiologists. However, existing works focus mainly on the chest area due to the limited availability of public datasets for other regions. Besides, they often rely on naive data-driven approaches, \eg, a basic encoder-decoder framework with captioning loss, which limits their ability to recognise complex patterns across diverse anatomical regions. To address these issues, we propose X-RGen, a radiologist-minded report generation framework across six anatomical regions. In X-RGen, we seek to mimic the behaviour of human radiologists, breaking them down into four principal phases: 1) initial observation, 2) cross-region analysis, 3) medical interpretation, and 4) report formation. Firstly, we adopt an image encoder for feature extraction, akin to a radiologist's preliminary review. Secondly, we enhance the recognition capacity of the image encoder by analysing images and reports across various regions, mimicking how radiologists gain their experience and improve their professional ability from past cases. Thirdly, just as radiologists apply their expertise to interpret radiology images, we introduce radiological knowledge of multiple anatomical regions to further analyse the features from a clinical perspective. Lastly, we generate reports based on the medical-aware features using a typical auto-regressive text decoder. Both natural language generation (NLG) and clinical efficacy metrics show the effectiveness of X-RGen on six X-ray datasets. Our code and checkpoints are available at: \href{https://github.com/YtongXie/X-RGen}{https://github.com/YtongXie/X-RGen}.

  \blfootnote{$^*$ Equal contributions. $^\dagger$ Corresponding author.}

\end{abstract}

\section{Introduction}
\label{sec:intro}

The tasks of interpreting radiology images and producing reports are both arduous and prone to errors. To reduce this burden, automatic report generation systems can provide candidate reports for radiologists to verify. Besides, these systems can leverage data-hungry machine learning paradigms by learning directly from free-text reports, which is a significant advantage compared to other medical image analysis applications (\eg, medical image segmentation~\cite{isensee2021nnu,lee20223d,zhang2021dodnet,wang2022multi}) that often rely on large amounts of quality annotations.

Radiologists commonly write reports based on radiology images covering different body parts.
Despite notable progress, existing report generation works~\cite{li2023dynamic,liu2021exploring,liu2021auto,ma2021contrastive,wang2023metransformer,zhang2020radiology,yan2023attributed} have primarily focused on the chest, a limitation stemming from the scarcity of publicly available datasets for other anatomical regions. This narrow focus hampers the broader clinical utility of these systems.
Besides, as they are designed following the typical single-dataset training-and-testing paradigm, they inevitably suffer from severe performance drop issues, when these generation models are directly deployed to another dataset \wrt various body regions.
By contrast, learning across various anatomical regions can potentially uncover underlying commonalities in medical images, \eg, the fracture in the wrist, shoulder, knee and other parts; or overlapping areas in chest and abdomen X-ray images.
Thus, it is crucial to design a report generation framework capable of covering multiple anatomical regions.

\begin{figure}[t]
    \centering
    \includegraphics[width=1.0\linewidth]{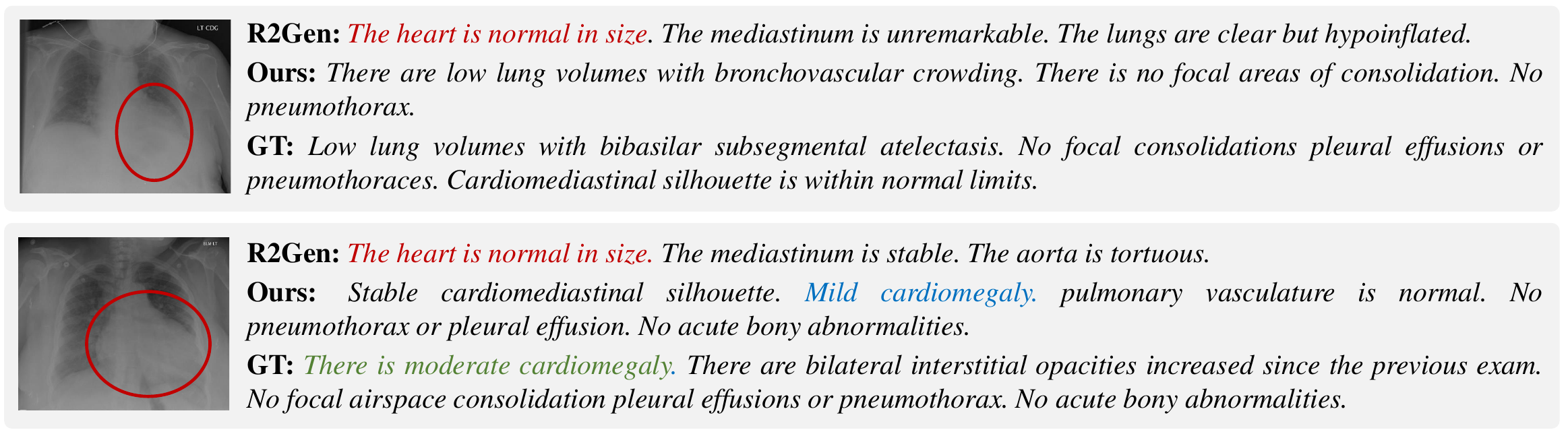}
    \vspace{-15pt}
    \caption{Reports written by radiologists vs. existing models (\eg, R2Gen~\cite{chen2020generating} trained on our merged dataset), and our X-RGen. We observe that R2Gen remembers some commonly used descriptions (highlighted in \textcolor{red}{red}) regardless of the semantic alignment with images, \eg, the correct diagnosis is \textit{``there is moderate cardiomegaly''} (highlighted in \textcolor{Green}{green}) while R2Gen keeps \textit{``the heart is normal in size''} (highlighted in \textcolor{red}{red}).}
    \vspace{-5pt}
    \label{fig:example}
\end{figure}

Technically, the heavy reliance on naive data-driven methods, such as basic encoder-decoder frameworks only with simple captioning loss, curtails their capability to identify complex medical patterns.
In this way, not all the generated reports are semantically consistent with the images as the model tends to remember an ``average'' version that contains the frequently occurring words and phrases present in the training corpus~\cite{chen2022learning} (see Figure~\ref{fig:example}).
However, in radiology reports, there are many rare but critically important medical terminology vital for diagnosis.
Thus, the challenge lies in designing a model that not only captures the typical data patterns but also recognises and accurately incorporates critical, albeit infrequent, medical terminology into radiology reports, ensuring a high degree of semantic consistency and diagnostic relevance.

\begin{figure}[t]
    \centering
    \begin{subfigure}[t]{0.55\linewidth}
        \centering
        \includegraphics[width=1.0\linewidth]{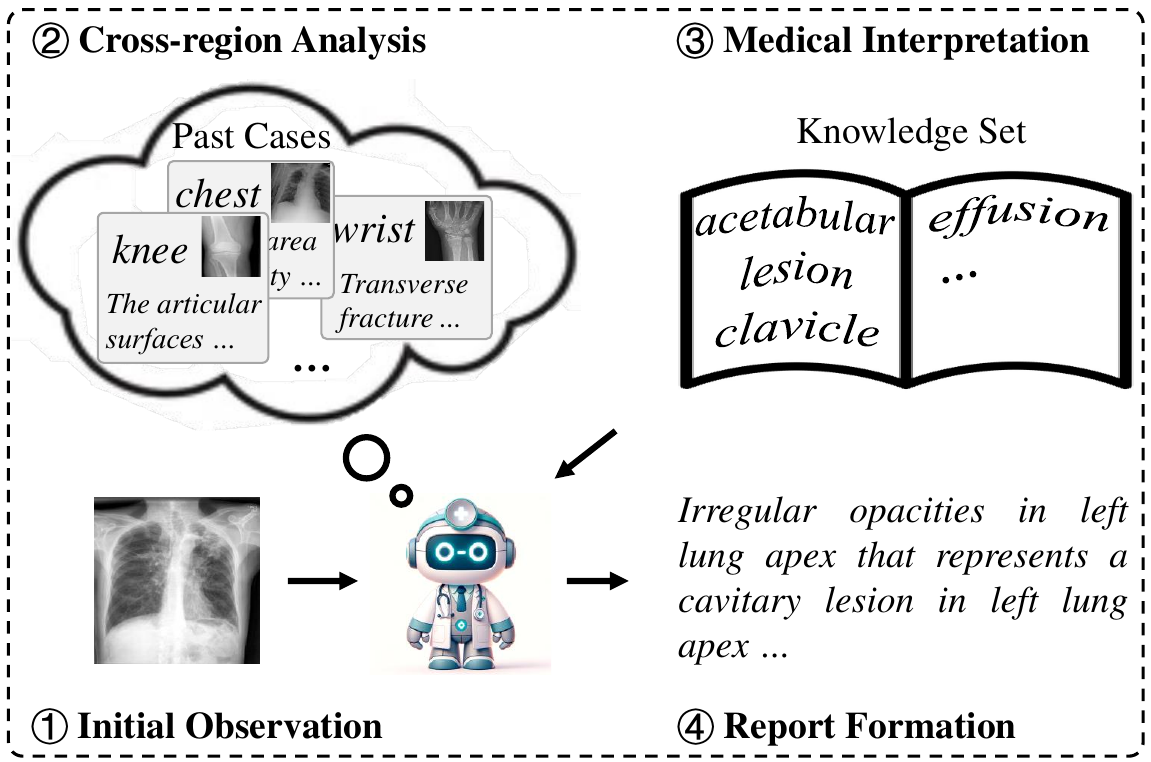}
        \caption{Intuitive understanding of our X-RGen.}
        \label{fig:intuitive}
    \end{subfigure}
    \begin{subfigure}[t]{0.43\linewidth}
        \centering
        \includegraphics[width=1.0\linewidth]{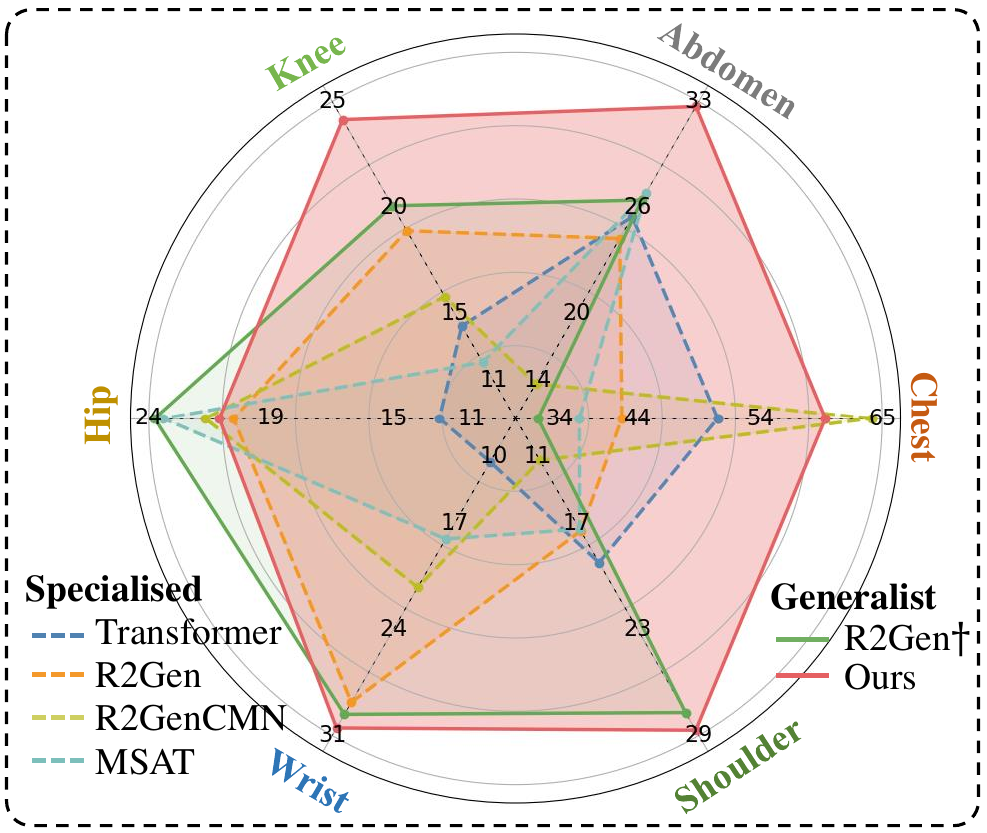}
        \caption{Results on CIDEr.}
        \label{fig:circle_cider}
    \end{subfigure}
    \vspace{-5pt}
    \caption{(a) X-RGen mimics the behaviour of how human radiologists write reports. (b) We calculate CIDEr for both specialised and generalist models on different datasets.}
\end{figure}

To address these issues, we propose a radiologist-minded framework for generating radiology reports across diverse anatomical regions, named X-RGen.
It covers six body parts: chest, shoulder, hip, knee, abdomen, and wrist.
As shown in Figure~\ref{fig:intuitive}, our X-RGen closely emulates the behaviour of human radiologists, which we have distilled into four key phases: 1) initial observation, 2) cross-region analysis, 3) medical interpretation, and 4) report formation.
Firstly, akin to a radiologist's initial assessment of medical images, X-RGen adopts an image encoder to identify crucial features within radiology images in the \textbf{initial observation} phase.
Then, much like how radiologists deepen their understanding of a patient's health by drawing upon their extensive experience with past cases, X-RGen aims to enhance the recognition ability of the image encoder by leveraging comparisons across images and reports from various anatomical regions in the \textbf{cross-region analysis} phase.
After that, mirroring radiologists' use of their expertise for image interpretation, our model similarly employs pre-defined radiological knowledge to conduct an in-depth clinical analysis of the enhanced features for \textbf{medical interpretation}. In this way, the model would pay more attention to medical-relevant terms, even those infrequent in the training corpus.
%
Finally, in the \textbf{report formation} phase, X-RGen compiles these insights into coherent and detailed reports.

We conduct experiments on a merged dataset, covering six different anatomical regions, \ie, chest, abdomen, knee, hip, wrist and shoulder. For a fair comparison, we include chest images from a widely used public dataset -- IU-Xray~\cite{demner2016preparing}. For the other five regions, we use our private data.
To evaluate the performance, we apply the natural language generation metrics (BLEU~\cite{papineni2002bleu} and CIDEr~\cite{vedantam2015cider}) and clinical efficacy metrics (recall and F1 score \cite{liu2019clinically}).
The results (see Figure~\ref{fig:circle_cider}) show the superiority of X-RGen compared with both specialised (trained on each single dataset) and generalist models (trained on the merged dataset).

In summary, our contributions include:
\begin{itemize}
    \item We propose X-RGen, a framework inspired by the behaviour of radiologists for generating reports across various anatomical regions. This framework contains four main phases: initial observation, cross-region analysis, medical interpretation, and report formation.
    \item We enhance image recognition through cross-region analysis (CA), improving alignment between images and reports across anatomical areas. In medical interpretation (MI), we integrate radiology-specific knowledge, alleviating the ignoring of rare yet crucial terms during report generation.
    \item We verify the superiority of our X-RGen on seven datasets \wrt different anatomical regions. The experimental results on both NLG and clinical efficacy metrics demonstrate the effectiveness of the proposed X-RGen.
\end{itemize}

\section{Related Works}

\subsubsection{Image Captioning}
Natural image captioning~\cite{anderson2018bottom,huang2019attention,vinyals2015show,xu2015show} seeks to automatically generate descriptive captions for a given image, garnering significant interest from researchers~\cite{lecun2015deep}.
Many methods~\cite{cornia2020meshed,lu2017knowing,pan2020x,rennie2017self} have been proposed, leading to significant advancements in the state-of-the-art.
The typical image captioning models~\cite{karpathy2015deep,vinyals2015show} mainly contain two components: a CNN-based image encoder and an RNN-based decoder for generating captions.
Several studies~\cite{huang2019attention,zhou2020unified} have incorporated the attention mechanism~\cite{vaswani2017attention} into the diagram, encouraging the models to pay greater attention to the highlighted regions.
However, radiology report generation requires specialised knowledge of medical imaging and terminology, while natural image captioning is more general in nature.

\subsubsection{Radiology Report Generation}
%
Radiology report generation focuses on medical imaging data to produce detailed and accurate reports that encapsulate findings, interpretations, and diagnoses from medical images.
Previous  works~\cite{jing2017automatic,wang2021self,xue2018multimodal,yuan2019automatic} employ a hierarchical LSTM for the long paragraph generation in medical reports.
To further enhance performance, several studies~\cite{chen2022cross,chen2020generating,li2022cross,wang2022automated,wang2023metransformer,hou2023recap} adopt a Transformer as the report decoder, leading to notable improvements in results.
Moreover, to capture the radiology terminologies and their semantic relationships, recent works~\cite{li2023dynamic,liu2021exploring,liu2021auto,zhang2020radiology} explore the incorporation of knowledge graphs as inputs or optimisation constraints (\eg, classification labels) in the generation process.
However, these models are designed based on a single-dataset training-testing paradigm, while radiologists often write reports according to radiology images \wrt various body regions, including chest, abdomen, \etc.
When these models are applied directly to another dataset that contains different body regions, they often encounter significant performance degradation issues.

While other knowledge-based models like~\cite{liu2021exploring,yang2022knowledge} develop a knowledge graph, the relations (edges) between topics (nodes) cannot be updated during training, which limits its effectiveness for exploiting implicit relationships. Besides, this graph focuses on chest X-rays only, restricting its applicability to other anatomical regions.
Thus, we tend to reorganise the topics in our knowledge set such that they cover the medical terminologies relevant to a broad range of body regions without predefined and fixed relations, where the relations are learnable.

\begin{figure*}[t]
    \centering
    \includegraphics[width=1.0\linewidth]{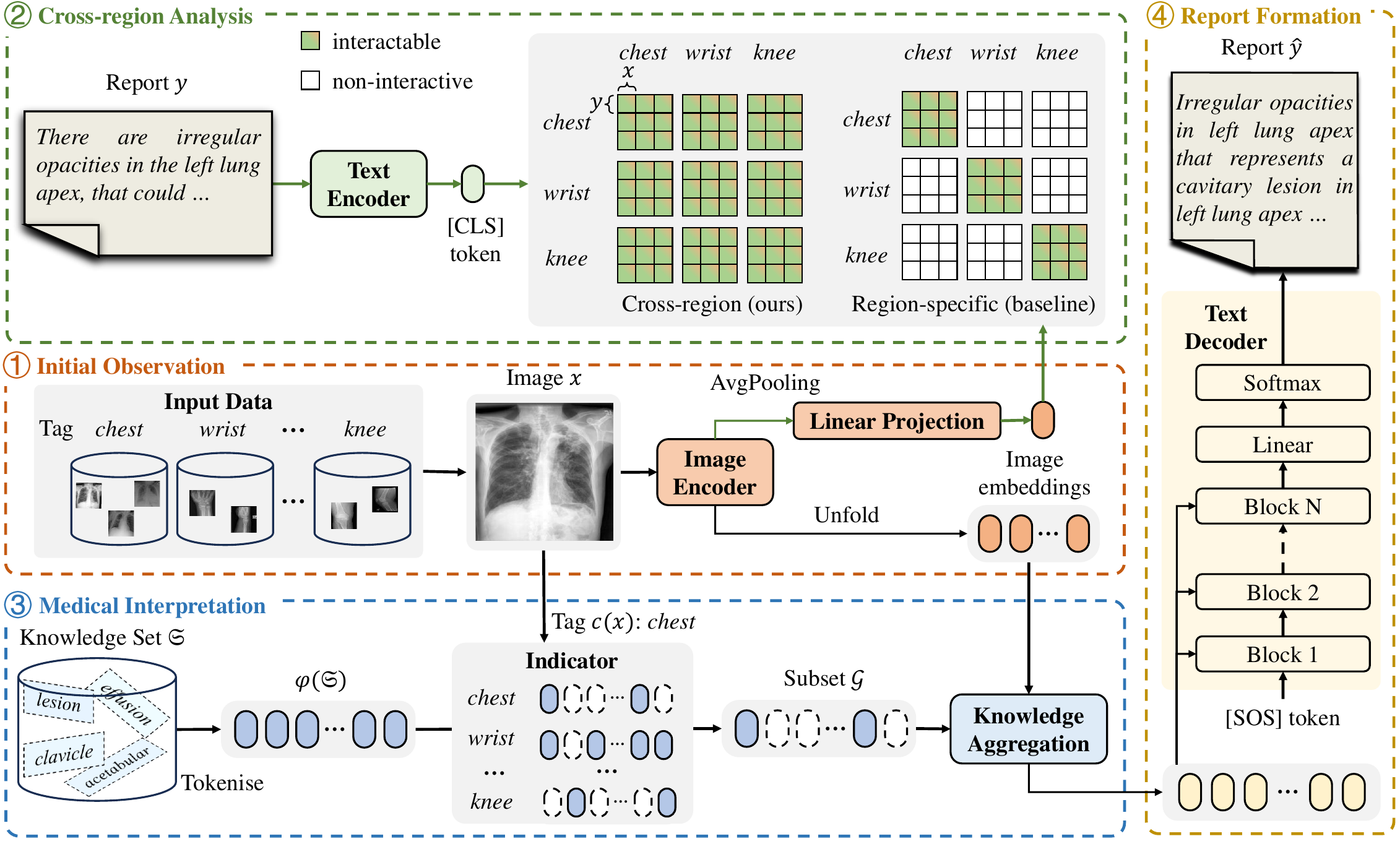}
    \caption{Overall of X-RGen. We decompose the framework into four phases: 1) initial observation, 2) cross-region analysis, 3) medical interpretation, and 4) report formation. 
    Specifically, starting with an image encoder to extract visual features, the model then enhances recognition by interacting with cross-region data. Next, it applies radiological knowledge for further medical-aware analysis, and finally, generates reports based on the enhanced and medical-aware features. 
    Note that the second phase (\ie, cross-region analysis, \textcolor{Green}{green} arrows) is only for training and will be removed in inference.
    }
    \label{fig:overall}
\end{figure*}

\section{Method}

Our X-RGen (see Figure~\ref{fig:overall}) contains four phases: 1) initial observation, 2) cross-region analysis, 3) medical interpretation and 4) report formation.
We first use an image encoder $f$ for feature extraction from images $x$. Then, we boost the image feature $\mathcal{O}$ to $\mathcal{\tilde{O}}$ during training by improving recognition ability through cross-region data interaction. Subsequently, we integrate radiological knowledge for a deeper medical interpretation and generate medically enhanced features $\mathcal{Z}$. Last, we yield a radiology report $y$ from $\mathcal{Z}$. Notably, the cross-region analysis phase (indicated by green arrows in Figure~\ref{fig:overall}) is excluded in inference.

\subsection{Initial Observation}

\qi{We introduce a CNN-based image encoder $f$ to simulate the initial examination phase of radiologists analysing radiology images. This encoder processes the input image $x$ using convolutional layers to extract feature maps, which capture crucial diagnostic information. 
The extracted features are unfolded through a \texttt{Unfold} operation, creating a set of feature embeddings.
This CNN architecture dynamically focuses on important image details, mirroring a radiologist’s method of identifying key features.}
%
Mathematically,
\begin{equation}
    \begin{aligned}
        \mathcal{O} = \texttt{Unfold}(f(x)),
    \end{aligned}
    \label{eq:patchify}
\end{equation}
\qi{where $\mathcal{O}$ is the visual embeddings, which can be defined as $\mathcal{O}=\{o_1, o_2, ..., o_n\}$.}
%
By simulating the initial observation phase, the image encoder can effectively capture and prioritise the most relevant features within the radiology images before proceeding to more detailed analysis and interpretation phases. This approach ensures that the encoder, much like a radiologist, establishes a global understanding of the image, setting a solid foundation for accurate medical interpretation and subsequent report generation.

\subsection{Cross-region Analysis}\label{sec:cross_region_analysis}

With the above observation, radiologists develop a further understanding of the patient's health status based on their experience, which is learned from reviewing and analysing numerous past cases. Similarly, in our cross-region analysis phase, we seek to improve the recognition ability of the image encoder $f$ by analysing images and reports from different anatomical regions.

\subsubsection{Unified Representation}

\qi{We first adopt the image encoder $f(x)$
to extract visual features from multi-region images, where $x$ represents images from various anatomical regions. Then, the extracted features are summarised through an average pooling (\texttt{AvgPooling}) followed by a linear projection layer\footnote{For simplicity, we represent the whole process as $\zeta$, including both \texttt{AvgPooling} and linear projection.}, creating a comprehensive image embedding with a specific dimension.}
For reports, we use a Transformer-based text encoder $g$ to process the corresponding reports $y$, \ie, $\mathcal{W} = g(y)$, where $\mathcal{W}=\{w_1, w_2, ..., w_m, w_{\texttt{[CLS]}}\}$ is the set of word tokens. 

\subsubsection{Enhancing Recognition with Cross-region Learning}

For a more comprehensive understanding of the human body, we seek to enable models to consider how different anatomical regions relate to each other.
From Figure~\ref{fig:overall}, previous region-specific works~\cite{li2023dynamic,jin2023promptmrg} focus only on alignments among images and reports within the same anatomical region. Unlike these, we adapt the learning objective for cross-region analysis, allowing for interactions across different anatomical regions.
\qi{We encode images and reports with $\zeta(f(x))$ and $g(y)$, respectively, into the shared space.} Formally, the cross-region learning objective can be defined as
%
\begin{equation}
    \begin{aligned}
    & \mathcal{L}_{i2r} = -\frac{1}{|\mathcal{B}|}\sum_{i=1}^{|\mathcal{B}|}\log \frac{\exp\Big(\sigma\big(\zeta(f(x_i)), g(y_i)\big)\Big)}{\sum^{|\mathcal{B}|}_{j=1,~i \neq j} \exp\Big(\sigma\big(\zeta(f(x_i)), g(y_j)\big)\Big)}, \\
    & \mathcal{L}_{r2i} = -\frac{1}{|\mathcal{B}|}\sum_{i=1}^{|\mathcal{B}|}\log \frac{\exp\Big(\sigma\big(g(y_i), \zeta(f(x_i))\big)\Big)}{\sum^{|\mathcal{B}|}_{j=1,~i \neq j} \exp\Big(\sigma\big(g(y_i), \zeta(f(x_j))\big)\Big)}, \\
    & \mathcal{L}_{\text{x}} = \frac{1}{2}(\mathcal{L}_{i2r} + \mathcal{L}_{r2i}),
    \end{aligned}
    \label{eq:cross_region}
\end{equation}
where $\sigma$ is the similarity function that calculates the cosine similarity between \qi{$\zeta(f(x))$} and $w_{\texttt{[CLS]}}\in\mathcal{W}=g(y)$. The sum in the denominator runs over all image-report pairs $(x, y)$ within the mini-batch $\mathcal{B}$.
Notably, the pairs selected within each mini-batch span multiple anatomical regions, which ensures diversity and holistic learning across different body regions.
In this way, the model gains a deeper, more generalised insight into the semantics across varied regions, improving its recognition capacity.
%
In general, the image embeddings before and after enhancement can be represented as
\begin{equation}
    \begin{aligned}
    \mathcal{T} & := \mathcal{O} \rightarrow \mathcal{\tilde{O}} \\
    & :=\{o_1, o_2, ..., o_n\} \rightarrow \{\tilde{o}_1, \tilde{o}_2, ..., \tilde{o}_n\},
    \end{aligned}
    \label{eq:cross_region_learning}
\end{equation}
where $\mathcal{T}$ refers to the process of our cross-region analysis during training. \qi{$\mathcal{\tilde{O}}$ is the enhanced image embeddings from the input image $x$.}



\subsection{Medical Interpretation}\label{sec:medical_interpretation}

\subsubsection{Building General Radiological Knowledge Set}

We seek to construct a general knowledge set $\mathfrak{S}$ that covers the most common abnormalities or findings in the radiology reports. For convenience, we call each item in this knowledge set a topic like~\cite{liu2021exploring,zhang2020radiology}.
While they develop a knowledge graph, the relations (edges) between topics (nodes) cannot be updated during training, which limits its effectiveness for exploiting implicit relationships. Furthermore, this knowledge graph focuses on chest X-rays only, which restricts its applicability to other body parts and broader use cases.
Thus, we have reorganised the topics in our knowledge set such that they cover the medical terminologies relevant to a broad range of body parts without pre-defined and fixed relations.

To better build a generic knowledge base, following~\cite{willemink2020preparing}, we adopt one of the most useful natural language processing methods, called topic modelling, on the database, that seeks to characterise the knowledge for each body part with a series of topics namely $\mathcal{G}$.
%
%
Specifically, we first use spacy~\cite{neumann-etal-2019-scispacy}, currently the most popular entity detection tool, to extract the medical entities and obtain the $50$ most frequent words for each body part. Then, $20\sim30$ most critical words are further filtered by radiologists to create the existing knowledge base.
Mathematically, the general set $\mathfrak{S}$ can be defined as $\mathfrak{S}=\mathcal{G}_1\cup\mathcal{G}_2\cup...$.
Due to the page limit, we put the details of the knowledge base in the supplementary.

\subsubsection{Region-aware Knowledge Selection}

Understanding the context and clinical nuances in radiology reports requires deep medical knowledge and expertise. Models may lack the comprehensive understanding needed to generate reports that incorporate relevant clinical information, leading to inaccuracies or missing crucial details.
To address this, we introduce a condition signal $c$ to selectively activate a set of topics based on the anatomical regions associated with the input image. 
By doing so, our model is able to filter out topics that are irrelevant to the specific region before conducting reasoning between the given image $x$ and the general knowledge set $\mathfrak{S}$.
Specifically, we devise an indication function $\mathbbm{1}(\cdot)$, which enables the selection of the topics in $\mathfrak{S}$.
Formally,
\begin{equation}
    \begin{aligned}
    \mathcal{G}=\mathbbm{1}(\varphi(\mathfrak{S})|c(x)),
    \end{aligned}
    \label{eq:indicator}
\end{equation}
where $\varphi$ is the pre-trained tokeniser for word embeddings while $\mathcal{G}$ is the subset of the general knowledge $\mathfrak{S}$ (\ie, $\mathcal{G}\subseteq\mathfrak{S}$), denoting the selected topics.
Here, $c(x)$ is a tag of the body region the given image $x$ belongs to (\eg, ``\textit{chest}''), which is manually predefined in advance\footnote{In clinical practice, since doctors specify which specific body part to image before taking medical images, the corresponding part naturally has a tag.}.
For instance, if the tag of an input image is predefined as ``\textit{chest}'', we choose a specific set of topics -- namely, ``\textit{airspace disease, atelectasis, calcinosis, cardiomegaly, cicatrix, edema, effusion, emphysema, fractures, hernia, hypoinflation, lesion, medical device, normal, opacity, other, pneumonia, pneumothorax, scoliosis, thickening}'' -- to represent the knowledge base for this image.
Notably, the topics selected for different samples belonging to the same body region are identical.
In our work, we define six different tags, including \textit{chest}, \textit{abdomen}, \textit{knee}, \textit{hip}, \textit{wrist}, and \textit{shoulder}.

\subsubsection{Intra-region Knowledge Aggregation}

For knowledge aggregation, our idea is to design a learnable aggregation model $\pi$ with a capacity for cross-modal reasoning, allowing it to determine the most relevant topics between the knowledge set $\mathcal{G}$ and the enhanced embeddings of image $\mathcal{\tilde{O}}$.
A straightforward way is adopting scaled dot product attention, which enables the topics and images to interact with one another.
Concretely, we design a knowledge-image co-attention module using a $l$-layer Transformer. 
Formally,
\begin{equation}
    \begin{aligned}
    \mathcal{Z} = \pi(\mathcal{\tilde{O}}, \mathcal{G}) = \texttt{Transformer}([\mathcal{\tilde{O}}; \mathcal{G}]),
    \end{aligned}
    \label{eq:aggregation}
\end{equation}
Here, $[\cdot;\cdot]$ is the concatenation operation. $\mathcal{Z}$ is the set of aggregated embeddings, \qi{\ie, $\mathcal{Z}=\{z_1, z_2, ... , z_n, ..., z_{n+k}\}$}, where $k$ is the number of topics in $\mathcal{G}$.

\subsection{Report Formation}

\subsubsection{Text Decoder}

Based on the aggregated embeddings $\mathcal{Z}$, we adopt a Transformer-based text decoder, containing $N$ Transformer blocks, for generating the final report (see Figure~\ref{fig:overall}).
Concretely, the decoding process starts by feeding a special start token $\texttt{[SOS]}$ to our text decoder, along with positional embeddings. The decoder uses a self-attention mechanism to process the start token, positional embeddings, and aggregated embeddings.
The whole process can be defined as
\begin{equation}
    \hat{y} = h(\mathcal{Z}) = \arg \max\prod_{t=1}^T p(\hat{w}_t|\hat{w}_{i<t}, \mathcal{Z}),
    \label{eq:text_decoder}
\end{equation}
where $h$ refers to the text decoder. $\hat{y}$ is the generated report and $\hat{w}_t$ is the $t$-th predicted word, \ie, $\hat{w}_t \in \hat{y}$.
The decoder generates the report auto-regressively, attending to the aggregated tokens and previously generated words at each step. 
In each step, it applies a softmax function to predict the next word's probability distribution over the entire vocabulary.
The process is repeated until an end token is generated or a predefined maximum sequence length $T$ is reached.

\begin{algorithm}[t]
\small
\caption{Overall Algorithm for X-RGen.}
\label{alg:training_inference}
    \begin{algorithmic}[1]
    	\REQUIRE 
        Training triplets $\{x, c(x), y\}$ \wrt image, tag, and report; X-RGen with modules: image encoder $f$, \qi{project layer $\zeta$}, text encoder $g$, aggregation model $\pi$, text decoder $h$.
    	\STATE Construct general knowledge set $\mathfrak{S}$ across multiple anatomical regions.
        \STATE // \emph{Training}
        \WHILE{\textit{not convergent}}
            \STATE Extract image embeddings $\mathcal{O}$ from each input image $x$ with Eq.~(\ref{eq:patchify}).
            \STATE Boost $\mathcal{O}$ to $\mathcal{\tilde{O}}$ by updating $f$, $\zeta$ and $g$ using the objective in Eq.~(\ref{eq:cross_region}).
            \STATE Select region-aware knowledge $\mathcal{G}$ from $\mathfrak{S}$ according to tag $c(x)$ in Eq.~(\ref{eq:indicator}).
            \STATE Obtain medical-aware image tokens $\mathcal{Z}$ by aggregating $\mathcal{G}$ and $\mathcal{\tilde{O}}$ with Eq.~(\ref{eq:aggregation}).
            \STATE Generate report $\hat{y}$ from $\mathcal{Z}$ with Eq.~(\ref{eq:text_decoder}).
            \STATE Update $f$, $\zeta$, $g$, $\pi$, and $h$ by minimising the objective in Eq.~(\ref{eq:final_loss}).
        \ENDWHILE
        \STATE // \emph{Inference}
        \STATE Extract embeddings $\mathcal{\tilde{O}}$ from image $x$ using $f$ with \texttt{Unfold} operation in Eq.~(\ref{eq:patchify}).
        \STATE Select knowledge $\mathcal{G}$ by Eq.~(\ref{eq:indicator}) and then aggregate it with $\mathcal{\tilde{O}}$ by Eq.~(\ref{eq:aggregation}) to get $\mathcal{Z}$.
        \STATE Generate report $\hat{y}$ from $\mathcal{Z}$ with Eq.~(\ref{eq:text_decoder}). 
    \end{algorithmic}
\end{algorithm}

\subsection{Training and Inference}

\subsubsection{Overall Training Objective}

As shown in Algorithm~\ref{alg:training_inference}, our overall training objective contains a captioning loss $\mathcal{L}_{cap}$ and the cross-region loss $\mathcal{L}_{\text{x}}$\footnote{$\mathcal{L}_{\text{x}}$ is the same as Eq.~(\ref{eq:cross_region}) in Section~\ref{sec:cross_region_analysis}.}, \ie,
\begin{equation}
    \begin{aligned}
        \mathcal{L} = \mathcal{L}_{cap} + \lambda\mathcal{L}_{\text{x}},
    \end{aligned}
    \label{eq:final_loss}
\end{equation}
where $\lambda$ is a hyper-parameter to balance these two terms.
Typically, sequence generation models are trained using the autoregressive Teacher Forcing scheme, to maximise the probability of the ground-truth token $w_t$ given all previous ground-truth tokens $w_{i<t}$. The captioning loss function can be formulated as
\begin{equation}
    \begin{aligned}
        \mathcal{L}_{cap}(x,y) = -\log p(y|x) = \sum^{T}_{t=1} -\log p(w_t|w_{i<t}, x),
    \end{aligned}
    \label{eq:caption_loss}
\end{equation}
where $w_t$ is the $t$-th token in report $y$, and $T$ is the total number of words in $y$.

\subsubsection{Inference}

As shown in Algorithm~\ref{alg:training_inference}, given a radiology image $x$, 
\qi{we use the image encoder $f$ to extract image features and use \texttt{Unfold} operation to obtains a set of image embeddings $\mathcal{\tilde{O}}$.}
After that, we select region-relevant knowledge $\mathcal{G}$ from the general knowledge set $\mathfrak{S}$ based on tag $c(x)$ and then aggregate $\mathcal{G}$ with image embeddings $\mathcal{\tilde{O}}$ to obtain the medical-aware features $\mathcal{Z}$. Last, we generate the report $\hat{y}$ from $\mathcal{Z}$.

Note that in inference, instead of relying on previous ground-truth word tokens, we predict the next word token based on the tokens that have been previously predicted in an auto-regressive manner.
Besides, the cross-region analysis is dropped in inference since its primary role is to boost the image encoder's recognition capabilities during training through the cross-region alignment loss.

\section{Experiments and Results}

\subsection{Datasets}


In experiments, we construct a merged dataset that contains paired data \wrt~six anatomical regions, including chest, abdomen, knee, hip, wrist and shoulder.
%
%
%
Due to the lack of existing datasets, we collect private image-report pairs on all six anatomical regions.
Anonymous Human Research Ethics Committee provides ethics approval for private data used in this study.
For each region, we have $3,000$ patients and the ratio of train/val/test is 70\%/15\%/15\%.
%
%
Notably, for a fair comparison with previous works, we use chest pairs on IU-Xray~\cite{demner2016preparing}, a publicly recognised dataset, rather than our private ones. It consists of $3,955$ fully de-identified radiology reports, each paired with frontal and/or lateral chest X-ray images.
Following~\cite{chen2020generating,li2023dynamic}, we remove cases that contain only a single image and then divide the dataset into train, validation, and test sets with 2069/296/590 pairs, respectively.
We put examples in the supplementary.


\subsection{Evaluation Metrics and Implementation Details}

\subsubsection{Evaluation Metrics}
%
%

To assess the quality of generated reports, we adopt widely used natural language generation (NLG) metrics, \ie, BLEU (B1$\sim$B4)~\cite{papineni2002bleu}, ROUGE~\cite{lin2004rouge}, METEOR~\cite{banerjee2005meteor} and CIDEr~\cite{vedantam2015cider}.
We access the clinical efficacy of generated reports using recall and F1 score \cite{liu2019clinically} along with a CLIP-based metric, called CLIPScore~\cite{hessel2021clipscore}.
The CLIPScore\footnote{We use MedClip~\cite{wang2022medclip} instead of the original CLIP trained on the natural domain.} can assess whether the generated reports are semantically aligned with given images, even when they are different from the reference reports.

To demonstrate the enhanced capacity for semantic understanding offered by the image encoder, we undertake the linear classification probing evaluation using the CheXpert~\cite{irvin2019chexpert} dataset, which contains five individual binary labels: atelectasis, cardiomegaly, consolidation, edema, and pleural effusion. For this process, we fix the image encoder, which has been trained on our X-RGen, and exclusively train a randomly initialised linear classification head.

\subsubsection{Implementation Details}
\qi{We adopt ResNet101~\cite{he2016deep}, pre-trained on ImageNet~\cite{deng2009imagenet}, serving as image encoder.}
We use the tokeniser and text encoder from MedClip~\cite{wang2022medclip} to convert words to embeddings.
The knowledge aggregation module consists of a three-layer Transformer~\cite{dosovitskiy2020image}. 
We resize input images to $224\times224$, and limit the maximum epochs to 100 and use Adam~\cite{kingma2014adam} with a weight decay of 1e-4. 
We set the $\lambda$ to 1.0.
We put more details in the supplementary.

\begin{table*}[t]
  \centering
  \resizebox{1.0\linewidth}{!}
  {
    \begin{tabular}{l|cccccccccccc|cc}
    \toprule
          & \multicolumn{2}{c}{Chest} & \multicolumn{2}{c}{Abdomen} & \multicolumn{2}{c}{Knee} & \multicolumn{2}{c}{Hip} & \multicolumn{2}{c}{Wrist} & \multicolumn{2}{c}{Shoulder} & \multicolumn{2}{|c}{Ave} \\
          & B4 & CIDEr & B4 & CIDEr & B4 & CIDEr & B4 & CIDEr & B4 & CIDEr & B4 & CIDEr & B4 & CIDEr \\
    \midrule
    \multicolumn{15}{c}{specialized models} \\
    \midrule
    Transformer~\cite{vaswani2017attention} &   0.162	&   0.511	&   \underline{0.108}	&   0.261	&   0.079	&   0.151	&   0.077	&   0.137	&   0.086	&   0.129	&   0.088	&   0.192	&   0.100	&   0.230  \\
    R2Gen~\cite{chen2020generating} &   0.165	&   0.430	&   0.105	&   0.248	&   0.077	&   \underline{0.193}	&   0.082	&   0.210	&   \underline{0.093}	&   \underline{0.288}	&   0.082	&   0.174	&   0.101	&   0.257  \\
    R2GenCMN~\cite{chen2022cross} &   0.170	& \underline{0.641}  	&   0.102	&   0.161	&   0.083	&   0.164	&   0.083	&   0.220	&   0.087	&   0.212	&   0.082	&   0.134	&   0.101	&   0.255 \\
    MSAT~\cite{wang2022medical} &   0.171	&   0.394	&   0.105	&   \underline{0.275}	&   0.082	&   0.135	&   0.081	&   \underline{0.235}	&   0.081	&   0.180	&   0.080	&   0.173	&   0.100	&   0.232  \\
   DCL~\cite{li2023dynamic} &   0.163  &    0.586   &   -    &   -    &  -     &  -     &    -   &  -     &   -    &   -    &   -    &  -  &  -     &  -  \\
   METransformer~\cite{wang2023metransformer} &   \underline{0.172}    &    0.435   &  -     &   -    &   -    &    -   &    -   &   -    &    -   &    -   &   -    &  -  &    -   &   - \\
   X-RGen (ours) & 0.163	    & 0.609	    & 0.106	    & 0.196	    & \underline{0.087}	    & 0.175	    & \underline{0.086}	    & 0.192	    & 0.089	    & 0.243	    & \underline{0.088}	    & \underline{0.197}	    & \underline{0.103}	    & \underline{0.269}   \\
    \midrule
    \multicolumn{15}{c}{generalist models} \\
    \midrule
    R2Gen$^\dagger$ (bs=16) & 0.084	    & 0.289	    & 0.104	    & 0.280	    & 0.064	    & 0.154	    & 0.074	    & 0.203	    & 0.085	    & 0.217	    & 0.082	      & 0.186	& 0.082	 & 0.222 \\
    R2Gen$^\dagger$ (bs=96) & 0.147	    & 0.470	    & 0.097	    & 0.271	    & 0.075	    & 0.181	    & 0.080	    & 0.226	    & 0.084	    & 0.258	    & 0.095	    & 0.274	    & 0.096	    & 0.280 \\
    R2Gen$^\dagger$ (bs=192) & 0.114	    & 0.359	    & 0.100	    & 0.271	    & 0.089	    & 0.204	    & \textbf{0.086}	    & 0.238	        & \textbf{0.102}	    & 0.296	    & 0.096	    & 0.277	    & 0.098	    & 0.274    \\
    X-RGen (ours, bs=16) &  0.152	&  0.509	&  0.108	&  0.276	&  0.071	&  0.166	&  0.073	&  0.184	&  0.079	&  0.229	&  0.084	&  0.220	&  0.095	&  0.264 \\
    X-RGen (ours, bs=96) & 0.161	    & \textbf{0.700}	    & 0.110	    & 0.292	    & 0.077	    & 0.188	    & 0.084	    & \textbf{0.257}	    & 0.090	    & 0.255	    & \textbf{0.099}	    & 0.272	    & 0.104	    & 0.327    \\
    X-RGen (ours, bs=192) & \textbf{0.177}	    & 0.602	    & \textbf{0.118}	    & \textbf{0.327}	    & \textbf{0.093}	    & \textbf{0.242}	    & 0.076	    & 0.215	    & 0.097	    & \textbf{0.305}	    & 0.096	    & \textbf{0.287}	    & \textbf{0.110}	    & \textbf{0.330}  \\
    \bottomrule
    \end{tabular}%
    }
    \resizebox{1.0\linewidth}{!}
    {
    \begin{tabular}{l|cccccccccccc|cc}
    \toprule
          & \multicolumn{2}{c}{Chest} & \multicolumn{2}{c}{Abdomen} & \multicolumn{2}{c}{Knee} & \multicolumn{2}{c}{Hip} & \multicolumn{2}{c}{Wrist} & \multicolumn{2}{c}{Shoulder} & \multicolumn{2}{|c}{Ave} \\
          & F & R & F & R & F & R & F & R & F & R & F & R & F & R \\
    \midrule
    \multicolumn{15}{c}{specialized models} \\
    \midrule
    Transformer~\cite{vaswani2017attention} & 0.584  	& 0.624 	& 0.559  	& 0.546  	& 0.486  	& \underline{0.464}  	& 0.525  	& 0.481  	& 0.506  	& 0.453  	& 0.463  	& 0.420  	& 0.521    	& 0.498    \\
    R2Gen~\cite{chen2020generating} & 0.583  	& \underline{\textbf{0.655}}  	& 0.558  	& 0.554  	& 0.462  	& 0.389  	& 0.496  	& 0.427  	& 0.514  	& 0.479  	& \underline{0.520}  	& 0.468  	& 0.522   	& 0.495      \\
    R2GenCMN~\cite{chen2022cross} & 0.592  	& 0.645  	& 0.540  	& 0.505  	& 0.491  	& 0.437  	& \underline{0.528}  	& 0.501  	& 0.500  	& 0.427  	& 0.462  	& 0.387  	& 0.484   	& 0.519     \\
   X-RGen (ours) & \underline{0.593} 	& 0.642  	& \underline{0.565}  	& \underline{0.559}  	& \underline{0.497}  	& 0.460  	& 0.522  	& \underline{\textbf{0.502}}	  	& \underline{0.533}  	& \underline{0.506}  	& 0.508  	& \underline{0.474}  	& \underline{0.536}   	& \underline{0.524}       \\
    \midrule
    \multicolumn{15}{c}{generalist models} \\
    \midrule
    R2Gen$^\dagger$ (bs=192) & 0.589   	&  0.578 	&  0.561	& 0.549  	& 0.495  	& 0.443  	& 0.512  	& 0.496  	& 0.531  	& 0.496  	& 0.505  	& 0.479  	& 0.532   	& 0.507        \\
    X-RGen (ours, bs=192)  & \textbf{0.594}  	& 0.647 	& \textbf{0.580}   	& \textbf{0.565}  	& \textbf{0.501}  	& \textbf{0.467}  	& \textbf{0.529}  	& 0.499  	& \textbf{0.543}  	& \textbf{0.512}  	& \textbf{0.514}  	& \textbf{0.482}  	& \textbf{0.544}   	& \textbf{0.529}         \\
    \bottomrule
    \end{tabular}%
    }
    \vspace{5pt}
    \caption{Comparison of NLG metrics (upper: B4 and CIDEr) and clinical efficacy metrics (lower: F $\rightarrow$ F1 ; R $\rightarrow$ recall) with the recent specialised models on six datasets. $^\dagger$ means we optimise the model on our merged training dataset while the ``bs'' is the training batch size. All evaluations are conducted on the test set. A higher value means better performance. We highlight the best results on specialised models with \underline{underline} while the best results on all models (both specialised and generalist) with \textbf{bold}.}
  \vspace{-20pt}
  \label{tab:iu_nlg_ce}%
\end{table*}%

\subsection{Comparison with State-of-the-arts}

\subsubsection{Specialised Baselines}

We compare X-RGen with the existing report generation methods, including R2Gen~\cite{chen2020generating}, R2GenCMN~\cite{chen2022cross}, MSAT~\cite{wang2022medical}, DCL~\cite{li2023dynamic} and METransformer~\cite{wang2023metransformer}. Besides, we consider a widely used natural image captioning method  (\ie, Transformer~\cite{vaswani2017attention}) as another baseline.
First, we individually optimise our model and each baseline in a specialised training setting.
For a fair comparison, we adopt the batch size (bs) of 16, which is a commonly used setting in the report generation task\footnote{We also experiment with increasing the batch size of the baselines to improve their performance, but it only results in performance comparable to $\text{bs}=16$.}.
%
In Table~\ref{tab:iu_nlg_ce}, compared with specialised baselines, our X-RGen achieves superior results in both NLG (average B4 and CIDEr) and clinical efficacy metrics (average F1 and recall scores). This indicates that the radiologist-minded framework benefits even in the specialised setting.

\subsubsection{Generalist Baselines}

To further analyse the performance of X-RGen, we adapt specialised models into the joint training setting due to the lack of existing generalist baselines. Here, we use all the training data on different subsets for optimisation.
To mitigate the impact of different architectures, we select R2Gen~\cite{chen2020generating} as the baseline. The main difference between R2Gen and our base model lies in the text decoder, where R2Gen has an additional Relational Memory (RM) module while our model does not include it.
For a fair comparison, we adjust the batch size (bs) to match our setting. Specifically, we increase it from 16 to 96 and 192, which aligns with our own configuration, thereby mitigating the potential performance improvement attributed solely to the larger batch size.

Table~\ref{tab:iu_nlg_ce} shows that regardless of $\text{bs}=96$ or $192$, our X-RGen consistently outperforms R2Gen in terms of both average B4 and CIDEr scores, which demonstrates its effectiveness in generating accurate and high-quality radiology reports.
Moreover, R2Gen (generalist) has an $\sim9\%$ improvement in CIDEr (0.257 to 0.280) while achieving a comparable result in B4 (0.101 and 0.098) compared with R2Gen (specialised). This indicates the positive impact of using diverse and increased training data.
For our X-RGen, the generalist version achieves larger improvements in both CIDEr ($\sim22\%$: 0.269 to 0.330) and B4 ($\sim7\%$: 0.103 to 0.110) compared with the specialised counterpart.
A similar phenomenon also occurs in clinical efficacy metrics (\ie, average F1 and recall scores) in Table~\ref{tab:iu_nlg_ce}.
These results demonstrate that the gains in performance are not solely attributed to the dataset, but also due to the benefits provided by the proposed radiologist-minded framework.
We put more results in the supplementary.

\vspace{-3pt}

\subsection{Qualitative Evaluation}

\vspace{-2pt}

\begin{figure*}[t]
    \centering
    \includegraphics[width=1.0\linewidth]{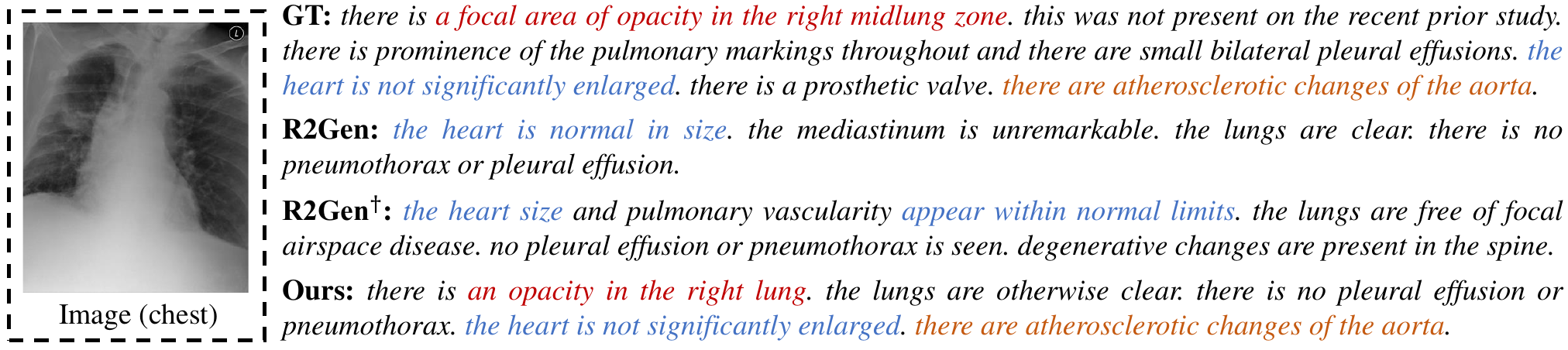}
    \vspace{-15pt}
    \caption{Reports generated by X-RGen (ours) and two baselines -- R2Gen and R2Gen$^\dagger$. R2Gen is trained on IU-Xray only while R2Gen$^\dagger$ optimised on our merged training set. 
    }
    \vspace{-15pt}
    \label{fig:visual_result}
\end{figure*}

In this part, we further assess the quality of reports generated by different methods, including our method and two baselines, \ie, R2Gen and R2Gen$^\dagger$, trained on IU-Xray (chest) only and our merged dataset, respectively.
In Figure~\ref{fig:visual_result}, we highlight the descriptions in different colours (red, blue and orange), which are semantically aligned with those in the ground-truth (GT) reports.
%
When considering the prominent area (\eg, the heart), all three models can provide (almost) accurate descriptions.
However, R2Gen still tends to generate the ``average'' descriptions like ``\textit{the heart is normal in size}'' while R2Gen$^\dagger$ shows improvement due to optimisation with a more diverse dataset.
Moreover, our X-RGen shows a more powerful capacity to generate descriptions that align semantically with the ground truth.
For instance, while the baselines fail to capture the details, our model accurately describes ``\textit{an opacity in the right lung}'', matching the GT description: ``\textit{a focal area of opacity in the right midlung zone}''.

\subsection{Ablation Study}

In this part, we evaluate the performance of our base model with and without the medical interpretation (MI) and cross-region analysis (CA) phases on Chest (IU-Xray). 
In Table~\ref{tab:ablation}, the results show that our base model with MI alone achieves better performance compared with the counterpart without it (\ie, B4: 0.129 $\rightarrow$ 0.156 while CIDEr: 0.426 $\rightarrow$ 0.537), which verifies the significance of the radiology-relevant knowledge in report generation task.
While MI$^*$, without the indicator $\mathbbm{1}(\cdot)$, can also achieve improved results compared to the base model (\eg, B4: 0.129 $\rightarrow$ 0.137), it is surpassed by MI (B4: 0.156). This highlights the importance of region-specific guidance and demonstrates the necessity of incorporating such guidance for better performance.
Finally, incorporating the CA phase further enhances the performance, resulting in the best scores for both B4 (0.161) and CIDEr (0.700). This demonstrates the contribution of the CA in improving the model performance by leveraging guidance from different modalities, including both images and reports.

\begin{table*}[t]
  \centering
  \resizebox{0.75\linewidth}{!}
  {
    \begin{tabular}{l|cccccccccccc|cc}
    \toprule
          & BLEU-1 & BLEU-2 & BLEU-3 & BLEU-4 & METEOR & ROUGE-L & CIDEr \\
    \midrule
    Base & 0.412	    & 0.255	    & 0.175	    & 0.129	    & 0.176     & 0.340	    & 0.426	     \\
    ~+MI$^*$ & 0.414	    & 0.263	    & 0.184	    & 0.137	    & 0.180     & 0.342	    & 0.474	     \\
    ~+MI & \textbf{0.457}	    & 0.284	    & 0.204	    & 0.156	    & 0.182     & 	0.349    & 0.537	    \\
    ~~~~~+CA & 0.454	&\textbf{0.290}	   &\textbf{0.210}	    &\textbf{0.161}	    &\textbf{0.187} &\textbf{0.361} & \textbf{0.700}	  \\
    \bottomrule
    \end{tabular}%
    }
    \vspace{5pt}
    \caption{Performance analysis on Chest (IU-Xray). ``Base'' contains only the initial observation and report formation phases. Both the medical interpretation (MI) phase and MI$^*$ are applied on top of the Base, where the MI$^*$ means MI without indicator $\mathbbm{1}(\cdot)$ in Eq.~(\ref{eq:indicator}). The cross-region analysis (CA) is applied only on top of the Base+MI.}
    \vspace{-20pt}
  \label{tab:ablation}%
\end{table*}%

\subsection{Discussions}

In this part, we explore how well our X-RGen aligns semantically between images and reports. We also assess the impact of our cross-region analysis (CA) and medical interpretation (MI) phases on this semantic alignment.
Besides, we evaluate the recognition capacity of our image encoder by linear probing on CheXpert.
Due to the page limit, we put more discussions in the supplementary, including the effect of different feature extractors, the impact of hyper-parameter $\lambda$, and whether feeding image tags into the model would cause information leakage.

\subsubsection{Semantic Alignment between Image and Report}

Besides the reference-based metrics like BLEU4, which may be influenced by semantically irrelevant factors (\eg, writing style~\cite{chen2022learning}), we seek to directly assess the semantic alignment between the input images and the generated reports.
Thus, we calculate a reference-free score, namely CLIPScore~\cite{hessel2021clipscore}, for R2Gen~\cite{chen2020generating} (trained on IU-Xray only), R2Gen$^\dagger$ (trained on the merged dataset) and our X-RGen.
Notably, as we use MedClip~\cite{wang2022medclip} in CLIPScore, which is pre-trained on chest X-ray datasets, we only evaluate this score on the IU-Xray (chest) dataset 
because it is open-sourced.
Besides, for a fair comparison, we set the batch size to 96 for both R2Gen$^\dagger$ (achieves the best results) and our model. For the specialised R2Gen, we keep the settings of the official code unchanged.
In Table~\ref{tab:CLIPScore}, our X-RGen outperforms R2Gen with a CLIPScore of 78.052, regardless of whether it is trained on IU-Xray only (77.670) or on our merged dataset (75.402).
%

Moreover, similarly to Table~\ref{tab:CLIPScore}, we use the CLIPScore to assess whether the model can generate more semantically aligned reports aided by two main phases (\ie, MI and CA).
Table~\ref{tab:report_image_alignment} reveals that the model with MI produces improved CLIPScore compared to the base counterpart (from 75.687 to 76.865), and incorporating the CA further enhances the performance, resulting in the best CLIPScore. 
It demonstrates the capability of our MI and CA phases in recognition enhancement, therefore generating more accurate reports.

\begin{table}[t]
    \centering
    \begin{subfloat}[Ours vs. R2Gen\label{tab:CLIPScore}]
    {
    \resizebox{0.25\linewidth}{!}
        {
        \begin{tabular}{l|c}
            \toprule
            & CLIPScore\\ 
            \midrule
            R2Gen~\cite{chen2020generating}  & 77.670 \\ 
            R2Gen$^\dagger$~\cite{chen2020generating} & 75.402 \\ 
            X-RGen (ours) & 78.052 \\ 
            \bottomrule
        \end{tabular}
        }
    }
    \end{subfloat}~~~
    \begin{subfloat}[Impact of MI and CA\label{tab:report_image_alignment}]
    {
    \resizebox{0.25\linewidth}{!}
        {
        \begin{tabular}{l|c}
            \toprule
              &  CLIPScore \\ 
            \midrule
            Base &  75.687  \\
            ~~+ MI &  76.865 \\
            ~~+ MI + CA &  78.052 \\
            \bottomrule
        \end{tabular}
        }
    }
    \end{subfloat}~~~
    \begin{subfloat}[Linear probing\label{tab:Linear}]
    {
    \resizebox{0.25\linewidth}{!}
        {
            \begin{tabular}{l|c}
            \toprule
            & AUC score \\ 
            \midrule
            R2Gen~\cite{chen2020generating} &    77.435    \\
            R2Gen$^\dagger$~\cite{chen2020generating} &   79.213  \\
            X-RGen w/o CA            & 80.405         \\
            X-RGen (ours)            &  81.252        \\
            \bottomrule
            \end{tabular}
        }
    }
    \end{subfloat}
    \caption{We assess (a) semantic alignment between images and reports on IU-Xray (chest), and (b) the effect of medical interpretation (MI) and cross-region analysis (CA) phases for alignment. (c) Linear probing on CheXpert to evaluate the recognition ability of the image encoder. $^\dagger$ means we optimise the model on our merged dataset.}
    \vspace{-20pt}
\end{table}

\subsubsection{Recognition Capacity of Image Encoder}

To further investigate the effect of our CA phase in recognition enhancement, we seek to directly test the recognition ability of our image encoder. To this end, we simply add a classification head on top of our image encoder (\ie, linear probing) and then evaluate the performance on a multi-label classification dataset -- CheXpert~\cite{irvin2019chexpert}.
%
In Table~\ref{tab:Linear}, our X-RGen obtains an 81.252 AUC score that outperforms both R2Gen (77.435) and R2Gen$^\dagger$ (79.213), which indicates the ability of our model to correctly recognise and classify different medical diseases within the input radiology images.
Moreover, we evaluate the performance of the X-RGen without incorporating CA during training. In this case, the AUC score decreases to 80.405, further demonstrating the effectiveness of CA in enhancing the recognition ability of our model.

\section{Conclusion}

In this paper, we propose X-RGen, a framework designed for automatic radiology report generation across multiple anatomical regions. Unlike previous works, our X-RGen follows the behaviour of human radiologists with four key phases: initial observation, cross-region analysis, medical interpretation, and report formation.
The experiments across six X-ray datasets demonstrate the superiority of our X-RGen.
Through this work, we hope to mark a step towards narrowing the gap between medical artificial intelligence and human radiologists, starting with a more radiologist-like diagnostic process for the report generation task.


%
%
\bibliographystyle{splncs04}
\bibliography{main}


\appendix

This document provides more discussions and experimental details to supplement the main submission. 
We organise the supplementary into the following sections.
\begin{itemize}
    \item In Section~\ref{sec:more_discussion}, we provide more discussions, including the effect of different feature extractors (Section~\ref{subsec:effect_feature_extractor}), the impact of hyper-parameter $\lambda$ (Section~\ref{subsec:lambda}), and whether feeding image tags into the model would cause information leakage (Section~\ref{subsec:leakage}).
    \item In Section~\ref{sec:example_private}, we show some examples on our private datasets.
    \item In Section~\ref{sec:knowledge_base}, we depict details of our general knowledge base.
    \item In Section~\ref{sec:implementation}, we provide more implementation details.
    \item In Section~\ref{sec:more_quantitative}, we show more quantitative results.
\end{itemize}

\section{More Discussions}\label{sec:more_discussion}

In this part, we provide more discussions, including the effect of different feature extractors in Section~\ref{subsec:effect_feature_extractor}, the impact of hyper-parameter $\lambda$ in Section~\ref{subsec:lambda}, and whether feeding image tags into the model would cause information leakage in Section~\ref{subsec:leakage}.

\subsection{Effect of Feature Extractors}\label{subsec:effect_feature_extractor}

In our X-RGen framework, the tokeniser for knowledge word embeddings is initialised using MedClip~\cite{wang2022medclip}. 
It, trained extensively on a vast corpus of clinical text, offers a robust choice for such feature extraction.
Meanwhile, within the cross-region analysis phase, the text encoder is initialised with MedClip as well. 
%
To empirically assess the contributions of the two medical-specific pre-training models, we modified our X-RGen, substituting these two pre-training feature extractors with a generic BERT pre-training~\cite{devlin2018bert}. For a fair comparison, we set all the batch sizes to 96. As shown in Table~\ref{tab:feature_extractor}, when initialised with this general-domain BERT, our X-RGen model experiences a performance degradation of approximately 22\% in CIDEr (declining from 0.324 to 0.302) and a 4\% decrease in B4 (from 0.104 to 0.100). 
The results demonstrate the significance of medical-specific initialisation. Nevertheless, even without it, our X-RGen significantly outperforms the base model. This suggests that the performance gains of the X-RGen framework are attributed not only to medical-aware initialisation but also to the cross-region analysis and medical interpretation phases we introduced.

\begin{table}[t]
    \centering
    \begin{subfloat}[Effect of different feature extractors\label{tab:feature_extractor}]
    {
    \resizebox{0.39\linewidth}{!}
        {
            \begin{tabular}{l|ccc}
            \toprule
             &  B4 & CIDEr \\ 
            \midrule
            Base  & 0.095 & 0.276 \\
            X-RGen with BERT init. &  0.100 & 0.302 \\
            X-RGen &  0.104 & 0.327 \\
            \bottomrule
            \end{tabular}
        }
    }
    \end{subfloat}~~~
    \begin{subfloat}[Impact of $\lambda$\label{tab:hyper}]
    {
    \resizebox{0.17\linewidth}{!}
        {
            \begin{tabular}{c|cc} 
            \toprule
             $\lambda$     & B4 & CIDEr \\ 
             \midrule
            0.5 & 	0.108	& 0.317  \\
            1.0 &   0.110	& 0.330  \\
            1.5	&   0.101	& 0.272   \\ 
            \bottomrule
            \end{tabular}%
        }
    }
    \end{subfloat}~~~
    \begin{subfloat}[Information leakage from tags\label{tab:info_leakage}]
    {
    \resizebox{0.35\linewidth}{!}
        {
            \begin{tabular}{l|ccc}
            \toprule
              &  B4 & CIDEr \\ 
            \midrule
            R2Gen~\cite{chen2020generating} &  0.096 & 0.280 \\
            R2Gen~\cite{chen2020generating} with tags &  0.097 & 0.284 \\
            \bottomrule
            \end{tabular}
        }
    }
    \end{subfloat}
    \caption{We test (a) the effect of different feature extractors. ``X-RGen with BERT init.'' means we initialise all text encoders in X-RGen with a generic BERT pre-training model; (b) Impact of hyper-parameter $\lambda$ in Eq.~(7); (c) whether feeding image tags $c(\cdot)$ into the model would cause information leakage. All results are on IU-Xray (chest).}
    \vspace{-15pt}
\end{table}

\subsection{Impact of Hyper-parameter $\lambda$ in Eq.~(7)}\label{subsec:lambda}

As shown in Table~\ref{tab:hyper}, when the value of $\lambda$ is small, such as $\lambda=0.5$, the performance of our X-RGen is suboptimal. The reason lies in the insufficient enhancement of the recognition across various anatomical regions and the semantic alignment between different modalities (\ie, images and reports).
As we increase the value of $\lambda$, the performance of X-RGen reaches its peak at $\lambda=1.0$. However, beyond that point, the performance starts to degrade.
To balance these two terms, we set the weighting parameter $\lambda$ to a value of $1.0$ in all our experiments.

\subsection{Risk of Information Leakage from Tag $c(x)$}\label{subsec:leakage}

%
To examine the absence of information leakage, we feed the tag $c(x)$ of each input image $x$ into the existing well-known R2Gen method and observe the impact of the performance.
As shown in Table~\ref{tab:info_leakage}, the inclusion of input tags does not lead to much-improved performance for R2Gen~\cite{chen2020generating} (\ie, B4: 0.096 $\rightarrow$ 0.097; CIDEr: 0.280 $\rightarrow$ 0.284). It implies that the presence of input tags $c(\cdot)$ does not result in information leakage. On the contrary, they can be considered as medical-related priors, but need a well-designed approach (\eg, the medical interpretation phase in our X-RGen) to unleash their inherent potential.

\section{Examples on Private Datasets}\label{sec:example_private}

\begin{figure}[t]
    \centering
    \includegraphics[width=1.0\linewidth]{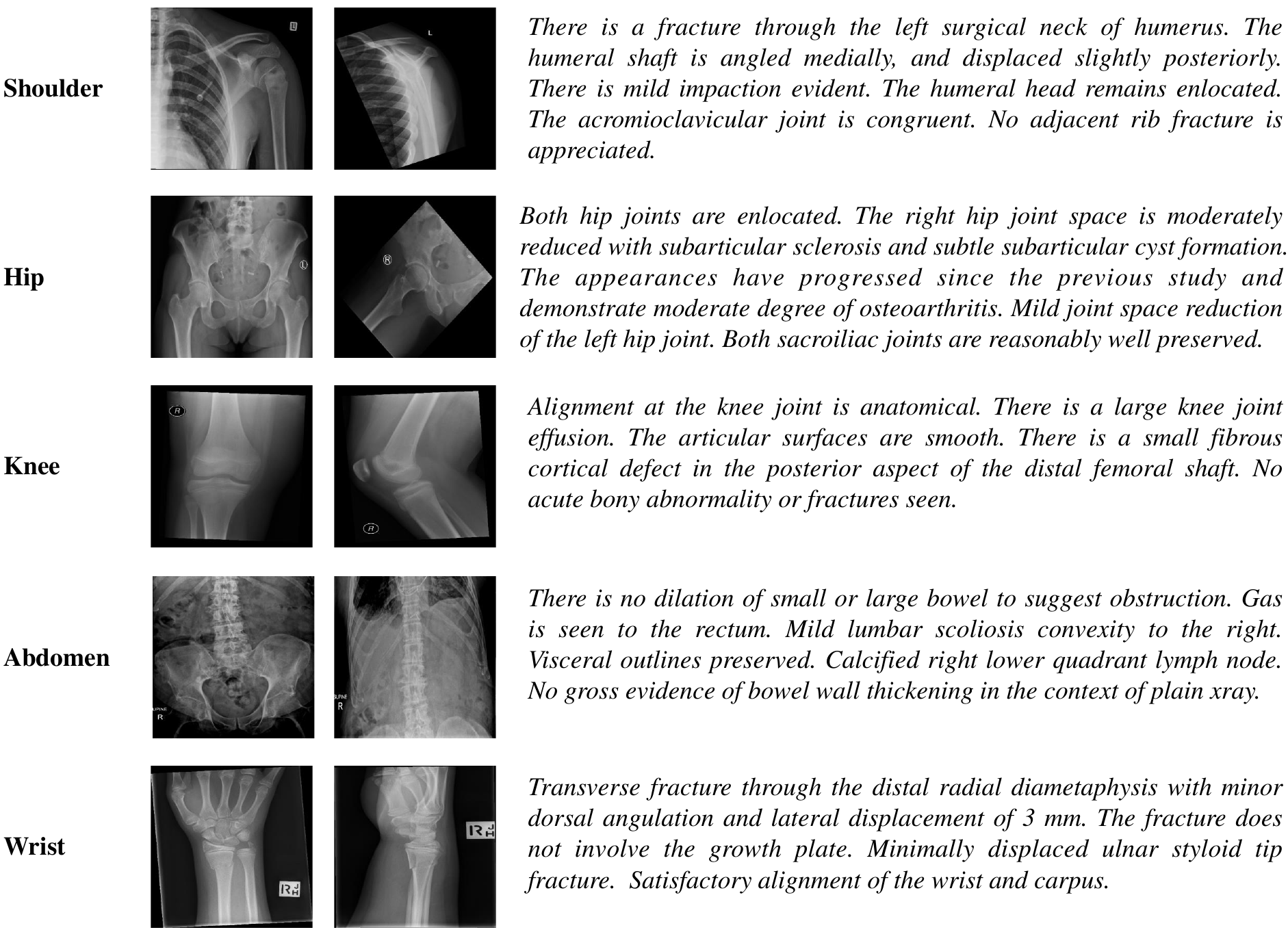}
    \caption{Examples on the private datasets. Each example contains a frontal image (first column) and another image (second column) with the corresponding radiology report.}
    \label{fig:supp_samples_private}
\end{figure}

In experiments, we construct a merged dataset that contains paired data \wrt~six anatomical regions, including chest, abdomen, knee, hip, wrist and shoulder.
Due to the lack of existing datasets, we collect private image-report pairs on all six anatomical regions.
Anonymous Human Research Ethics Committee provides ethics approval for private data used in this study.
For each region, we have $3,000$ patients and the ratio of train/val/test is 70\%/15\%/15\%.
Notably, for a fair comparison with previous works, we use chest pairs on IU-Xray~\cite{demner2016preparing}, a publicly recognised dataset, rather than our private ones. It consists of $3,955$ fully de-identified radiology reports, each paired with frontal and/or lateral chest X-ray images.
Following~\cite{chen2020generating,li2023dynamic}, we remove cases that contain only a single image and then divide the dataset into train, validation, and test sets with 2069/296/590 pairs, respectively.
Here, we provide some samples on the other five private datasets in Figure~\ref{fig:supp_samples_private}.

\section{Details of Knowledge Base}\label{sec:knowledge_base}

Here, we used different colours to highlight shared topics across the six anatomical regions. The results show that there are many topics commonly used, even across different regions. This finding indicates that our knowledge set has a relatively general scope.
Topics on our general knowledge set $\mathfrak{S}$ include:

\noindent\textit{\{abdomen, acetabular, acromioclavicular, acute, airspace disease, anatomical, angulation, atelectasis, bilateral, bone, bony, bowel, calcification, calcinosis, cardiomediastinal, cardiomegaly, carpal, cast, change, changes, cicatrix, clavicle, colon, compartment, complication, consolidation, contours, cuff, degenerative, dislocation, displacement, distal, dorsal, edema, effusion, emphysema, enlocated, evidence, faecal, femoral, femur, fracture, fractures, gas, glenohumeral, glenoid, head, healing, hernia, hip, humeral, humerus, hypoinflation, identified, inferior, intact, interval, joint, knee, lateral, lesion, limits, loading, loops, lucency, lumbar, lung, material, medical device, mild, moderate, nonspecific, normal, obstruction, opacity, other, patella, patellar, pelvic, pelvis, periprosthetic, plate, pleural, pneumonia, pneumothorax, projection, prosthesis, proximal, pubic, quadrant, radial, radio-carpal, radius, rectum, replacement, ring, sacroiliac, satisfactory, scaphoid, sclerosis, scoliosis, shoulder, situ, soft, space, stomach, styloid, subacromial, subdiaphragmatic, supine, suprapatellar, surgical, swelling, symphysis, thickening, tissue, tissues, transverse, tuberosity, ulnar, visualised, wrist\}}

Topics on each anatomical region namely $\mathcal{G}$ and we highlight the overlapped topics across different body parts in various colours:

\begin{itemize}

\item Chest = \{\textit{airspace disease, atelectasis, calcinosis, cardiomegaly, cicatrix, edema, \textcolor{red}{effusion}, emphysema, \textcolor{blue}{fractures}, hernia, hypoinflation, lesion, medical device, normal, opacity, other, pneumonia, pneumothorax, scoliosis, thickening}\}

\item Abdomen = \{\textit{abdomen, bowel, cardiomediastinal, colon, consolidation, contours, \textcolor{SpringGreen}{degenerative}, \textcolor{Emerald}{evidence}, faecal, \textcolor{Apricot}{gas}, limits, loading, loops, lumbar, lung, material, \textcolor{Purple}{moderate}, nonspecific, obstruction, pleural, projection, quadrant, rectum, stomach, subdiaphragmatic, supine, surgical, \textcolor{RawSienna}{tissue}}\}

\item Knee = \{\textit{\textcolor{CarnationPink}{acute}, \textcolor{Gray}{alignment}, \textcolor{Blue}{anatomical}, changes, compartment, complication, \textcolor{SpringGreen}{degenerative}, \textcolor{green}{dislocation}, \textcolor{red}{effusion}, \textcolor{Emerald}{evidence}, \textcolor{magenta}{femoral}, \textcolor{cyan}{fracture}, \textcolor{Apricot}{gas}, \textcolor{Yellow}{joint}, knee, \textcolor{BlueGreen}{lateral}, \textcolor{Periwinkle}{lucency}, \textcolor{Peach}{mild}, \textcolor{Purple}{moderate}, patella, patellar, prosthesis, \textcolor{Tan}{proximal}, replacement, satisfactory, situ, soft, suprapatellar, \textcolor{Mulberry}{swelling}, \textcolor{RawSienna}{tissue}, tissues}\}

\item Hip = \{\textit{acetabular, \textcolor{CarnationPink}{acute}, \textcolor{Gray}{alignment}, bilateral, bone, \textcolor{VioletRed}{bony}, \textcolor{SpringGreen}{degenerative}, enlocated, \textcolor{magenta}{femoral}, femur, \textcolor{cyan}{fracture}, \textcolor{blue}{fractures}, hip, \textcolor{YellowGreen}{identified}, \textcolor{Melon}{intact}, \textcolor{Yellow}{joint}, \textcolor{Periwinkle}{lucency}, \textcolor{Peach}{mild}, \textcolor{Purple}{moderate}, pelvic, pelvis, periprosthetic, \textcolor{Tan}{proximal}, pubic, ring, sacroiliac, sclerosis, symphysis}\}

\item Wrist = \{\textit{\textcolor{CarnationPink}{acute}, \textcolor{Gray}{alignment}, \textcolor{Blue}{anatomical}, angulation, \textcolor{VioletRed}{bony}, carpal, cast, \textcolor{SpringGreen}{degenerative}, displacement, distal, dorsal, \textcolor{cyan}{fracture}, healing, \textcolor{Melon}{intact}, interval, \textcolor{BlueGreen}{lateral}, \textcolor{Peach}{mild}, plate, radial, radio-carpal, radius, scaphoid, styloid, \textcolor{Mulberry}{swelling}, \textcolor{RawSienna}{tissue}, transverse, ulnar, wrist}\}

\item Shoulder = \{\textit{acromioclavicular, \textcolor{CarnationPink}{acute}, \textcolor{Gray}{alignment}, \textcolor{VioletRed}{bony}, calcification, change, clavicle, cuff, \textcolor{SpringGreen}{degenerative}, \textcolor{green}{dislocation}, \textcolor{cyan}{fracture}, \textcolor{blue}{fractures}, glenohumeral, glenoid, head, humeral, humerus, \textcolor{YellowGreen}{identified}, inferior, \textcolor{Melon}{intact}, \textcolor{Yellow}{joint}, \textcolor{BlueGreen}{lateral}, \textcolor{Tan}{proximal}, shoulder, space, subacromial, \textcolor{RawSienna}{tissue}, tuberosity, visualised}\}

\end{itemize}

\section{More Implementation Details}\label{sec:implementation}

Considering the domain disparity between medical and generic texts, we use the tokeniser and text encoder from MedClip~\cite{wang2022medclip} to embed the report. The knowledge aggregation network consists of a three-layer Transformer~\cite{dosovitskiy2020image}. \qi{For a fair comparison, following the setting of previous works, we configure the dimensions of input images to $224\times224$} and incorporate data augmentation techniques, such as random cropping and flipping, to expand the X-ray training dataset. We limit the maximum epochs to 100 and use the Adam optimiser~\cite{kingma2014adam} with a weight decay parameter of 1e-4. 
The learning rates are set at 5e-5 for the image encoder and 1e-4 for the remaining trainable parameters. Besides, based on the findings from our ablation study, we empirically set the hyper-parameter $\lambda$ to 1.0. Our experiments are conducted using A100 GPUs.

\section{More Quantitative Results}\label{sec:more_quantitative}

To assess the quality of the generated captions, we use four widely used NLG evaluation metrics, \ie, BLEU (B1$\sim$B4)~\cite{papineni2002bleu}, ROUGE~\cite{lin2004rouge}, METEOR~\cite{banerjee2005meteor} and CIDEr~\cite{vedantam2015cider}.
As shown in Table~\ref{tab:average}, we report the average scores of all the above evaluation metrics. The results exhibit that regardless of $\text{bs}=96$ or $192$, our X-RGen consistently outperforms R2Gen in terms of all the average scores (except for ROUGE-L), which demonstrates its effectiveness in generating accurate and high-quality radiology reports.
%
%
Specifically, when comparing R2Gen to our X-RGen in both the specialised and generalist settings, the improvements of R2Gen are $2.1\%$, $-0.4\%$, $-2.6\%$, $-2.9\%$, $5.6\%$, $-2.2\%$ and $8.9\%$ for BLEU-1, BLEU-2, BLEU-3, BLEU-4, METEOR, ROUGE-L and CIDEr, respectively\footnote{For a fair comparison, we compare the highest results for both R2Gen and ours.}. In contrast, our X-RGen achieves even larger improvements in these evaluation metrics about $8.3\%$, $7.4\%$, $6.7\%$, $6.8\%$, $6.9\%$, $-0.6\%$ and $22.7\%$ separately.
Moreover, we also report the values of all the evaluation metrics on these six datasets from Tables~\ref{tab:IU_Xray} to~\ref{tab:Shoulder}.

\begin{table*}[hbpt]
  \centering
  \caption{\textbf{Average} results on the six datasets compared with the recent specialised models.
  $^\dagger$ means we optimise the model on our merged training dataset while the ``bs'' is the training batch size. All evaluations are conducted on the test set, and a higher value indicates better performance.}
  \resizebox{1.0\linewidth}{!}
  {
    \begin{tabular}{l|ccccccc}
    \toprule
          & BLEU-1 (Ave) & BLEU-2 (Ave) & BLEU-3 (Ave) & BLEU-4 (Ave) & METEOR (Ave) & ROUGE-L (Ave) & CIDEr (Ave)  \\
    \midrule
    \multicolumn{8}{c}{specialised models} \\
    \midrule
    Transformer~\cite{vaswani2017attention} & 0.368 & 0.223 & 0.147 & 0.100 & 0.134 & 0.305 & 0.230 \\
    R2Gen~\cite{chen2020generating}& 0.374 & 0.229 & 0.149 & 0.101 & 0.141 & \textbf{0.312} & 0.257 \\
    R2GenCMN~\cite{chen2022cross} & 0.371 & 0.229 & 0.150 & 0.101 & 0.138 & 0.307 & 0.255 \\
    MSAT~\cite{wang2022medical} & 0.393 & 0.237 & 0.151 & 0.100 & 0.139 & 0.302 & 0.232 \\
    X-RGen (ours) & 0.370 & 0.227 & 0.150 & 0.103 & 0.144 & \textbf{0.312} & 0.269 \\
    \midrule
    \multicolumn{8}{c}{generalist models} \\
    \midrule
    R2Gen$^\dagger$ (bs=16) & 0.345 & 0.200 & 0.126 & 0.082 & 0.133 & 0.289 & 0.222  \\
    R2Gen$^\dagger$ (bs=96) & 0.382 & 0.228 & 0.145 & 0.096 & 0.149 & 0.301 & 0.280 \\
    R2Gen$^\dagger$ (bs=192)& 0.369 & 0.225 & 0.145 & 0.098 & 0.146 & 0.305 & 0.274  \\
     X-RGen (ours, bs=16) & 0.363 & 0.217 & 0.140 & 0.095 & 0.144 & 0.296 &  0.264 \\
     X-RGen (ours, bs=96) & 0.383 & 0.231 & 0.151 & 0.104 & 0.149 & 0.306 & 0.327   \\
     X-RGen (ours, bs=192) & \textbf{0.401} & \textbf{0.244} & \textbf{0.160} & \textbf{0.110} & \textbf{0.154} & 0.310 &  \textbf{0.330} \\
    \bottomrule
    \end{tabular}%
    }
  \label{tab:average}%
\end{table*}%

\begin{table*}[hbpt]
  \centering
  \caption{Comparison with the recent specialised models on Chest (IU-Xray).
  $^\dagger$ means we optimise the model on our merged training dataset while the ``bs'' is the training batch size. All evaluations are conducted on the test set, and a higher value indicates better performance.}
  \resizebox{0.8\linewidth}{!}
  {
    \begin{tabular}{l|ccccccc}
    \toprule
          & BLEU-1 & BLEU-2 & BLEU-3 & BLEU-4 & METEOR & ROUGE-L & CIDEr  \\
    \midrule
    \multicolumn{8}{c}{specialised models} \\
    \midrule
    Transformer~\cite{vaswani2017attention} &  0.459 & 0.298 & 0.215 & 0.162 & 0.188 & 0.362 & 0.511 \\
    R2Gen~\cite{chen2020generating} &  0.470 & 0.304 & 0.219 & 0.165 & 0.187 & 0.371 & 0.430 \\
    R2GenCMN~\cite{chen2022cross} & 0.475  & 0.309 & 0.222 & 0.170 & 0.191 & 0.375 & 0.641 \\
    MSAT~\cite{wang2022medical} & 0.481  & 0.316 & 0.226 & 0.171 & 0.190 & 0.372 & 0.394 \\
   DCL~\cite{li2023dynamic} &  - & - & - & 0.163 &  0.193 & \textbf{0.383} & 0.586   \\
   METransformer~\cite{wang2023metransformer} & \textbf{0.483}  & \textbf{0.322} & \textbf{0.228} & 0.172 & 0.192 & 0.380 &  0.435 \\
    X-RGen (ours) &0.441	&0.285	&0.208	&0.163	&0.184	&0.361	&0.609  \\
    \midrule
    \multicolumn{8}{c}{generalist models} \\
    \midrule
    R2Gen$^\dagger$ (bs=16) &0.306	&0.175	&0.117	&0.084	&0.134	&0.316	&0.289   \\
    R2Gen$^\dagger$ (bs=96) &0.433	&0.275	&0.196	&0.147	&0.184	&0.355	&0.470  \\
    R2Gen$^\dagger$ (bs=192) &0.349	&0.217	&0.153	&0.114	&0.154	&0.332	&0.359   \\
     X-RGen (ours, bs=16) &0.444	&0.287	&0.202	&0.152	&0.190	&0.365	&0.509   \\
     X-RGen (ours, bs=96) &0.454	&0.290	&0.210	&0.161	&0.187	&0.361	& \textbf{0.700}    \\
     X-RGen (ours, bs=192) &0.466	&0.306	&0.225	& \textbf{0.177} & \textbf{0.199} &0.367	&0.602   \\
    \bottomrule
    \end{tabular}%
    }
  \label{tab:IU_Xray}%
\end{table*}%

\begin{table*}[hbpt]
  \centering
  \caption{Comparison with the recent specialised models on Abdomen.
  $^\dagger$ means we optimise the model on our merged training dataset while the ``bs'' is the training batch size. All evaluations are conducted on the test set, and a higher value indicates better performance.}
  \resizebox{0.8\linewidth}{!}
  {
    \begin{tabular}{l|ccccccc}
    \toprule
          & BLEU-1 & BLEU-2 & BLEU-3 & BLEU-4 & METEOR & ROUGE-L & CIDEr  \\
    \midrule
    \multicolumn{8}{c}{specialised models} \\
    \midrule
    Transformer~\cite{vaswani2017attention} & 0.409 & 0.247	& 0.161 & 0.108 & 0.142 & 0.314 & 0.261  \\
    R2Gen~\cite{chen2020generating} & 0.389 & 0.241 &	0.156 &	0.105 &	0.143 &	0.309 &	0.248 \\
    R2GenCMN~\cite{chen2022cross} & 0.361 &	0.231 &	0.151 &	0.102& 0.135 & 0.310 &	0.161  \\
    MSAT~\cite{wang2022medical} & 0.410 & 0.246	& 0.157 & 0.105	& 0.140 & 0.286 & 0.275  \\
    X-RGen (ours) &0.373	&0.228	&0.154	&0.106	&0.137	&0.314	&0.196  \\
    \midrule
    \multicolumn{8}{c}{generalist models} \\
    \midrule
    R2Gen$^\dagger$ (bs=16) &0.386	&0.238	&0.154	&0.104	&0.144	&0.297	&0.280   \\
    R2Gen$^\dagger$ (bs=96) &0.407	&0.244	&0.150	&0.097	&0.155	&0.297	&0.271  \\
    R2Gen$^\dagger$ (bs=192) &0.397	&0.240	&0.151	&0.100	&0.153	&0.296	&0.271  \\
     X-RGen (ours, bs=16) &0.395	&0.243	&0.159	&0.108	&0.152	&0.305	&0.276  \\
     X-RGen (ours, bs=96) &0.409	&0.252	&0.162	&0.110	&0.159	&0.313	&0.292 \\
     X-RGen (ours, bs=192) & \textbf{0.432}	& \textbf{0.269} & \textbf{0.175} & \textbf{0.118} & \textbf{0.161} & \textbf{0.322} & \textbf{0.327} \\
    \bottomrule
    \end{tabular}%
    }
  \label{tab:Abdomen}%
\end{table*}%

\begin{table*}[hbpt]
  \centering
  \caption{Comparison with the recent specialised models on Knee.
  $^\dagger$ means we optimise the model on our merged training dataset while the ``bs'' is the training batch size. All evaluations are conducted on the test set, and a higher value indicates better performance.}
  \resizebox{0.8\linewidth}{!}
  {
    \begin{tabular}{l|ccccccc}
    \toprule
          & BLEU-1 & BLEU-2 & BLEU-3 & BLEU-4 & METEOR & ROUGE-L & CIDEr  \\
    \midrule
    \multicolumn{8}{c}{specialised models} \\
    \midrule
    Transformer~\cite{vaswani2017attention}  & 0.304 & 0.177 & 0.116 & 0.078 & 0.115 & 0.288 & 0.169  \\
    R2Gen~\cite{chen2020generating} & 0.308 & 0.191 & 0.121 & 0.077 & 0.130 & 0.300 & 0.193
     \\
    R2GenCMN~\cite{chen2022cross} & 0.329 & 0.201 &	0.130 &	0.083 &	0.120 &	0.284 &	0.164   \\
    MSAT~\cite{wang2022medical} & \textbf{0.366} & 0.203 & 0.128 & 0.082 & 0.134 & 0.282 & 0.135 \\
    X-RGen (ours) &0.339	&0.207	&0.133	&0.087	&0.135	&0.295	&0.175  \\
    \midrule
    \multicolumn{8}{c}{generalist models} \\
    \midrule
    R2Gen$^\dagger$ (bs=16) &0.321	&0.170	&0.100	&0.064	&0.119	&0.255	&0.154  \\
    R2Gen$^\dagger$ (bs=96) &0.343	&0.197	&0.120	&0.075	&0.134	&0.284	&0.181  \\
    R2Gen$^\dagger$ (bs=192) &0.333	&0.207	&0.134	&0.089	& \textbf{0.139} & \textbf{0.308} &0.204  \\
     X-RGen (ours, bs=16) &0.315	&0.180	&0.111	&0.071	&0.124	&0.276	&0.166  \\
     X-RGen (ours, bs=96) &0.331	&0.193	&0.120	&0.077	&0.130	&0.277	&0.188  \\
     X-RGen (ours, bs=192) & 0.359	& \textbf{0.219} & \textbf{0.141} & \textbf{0.093} & \textbf{0.139}	&0.291	& \textbf{0.242}  \\
    \bottomrule
    \end{tabular}%
    }
  \label{tab:Knee}%
\end{table*}%

\begin{table*}[hbpt]
  \centering
  \caption{Comparison with the recent specialised models on Hip.
  $^\dagger$ means we optimise the model on our merged training dataset while the ``bs'' is the training batch size. All evaluations are conducted on the test set, and a higher value indicates better performance.}
  \resizebox{0.8\linewidth}{!}
  {
    \begin{tabular}{l|ccccccc}
    \toprule
          & BLEU-1 & BLEU-2 & BLEU-3 & BLEU-4 & METEOR & ROUGE-L & CIDEr  \\
    \midrule
    \multicolumn{8}{c}{specialised models} \\
    \midrule
    Transformer~\cite{vaswani2017attention} & 0.334 & 0.193 & 0.118 & 0.077 & 0.116 &0.264 & 0.137 \\
    R2Gen~\cite{chen2020generating}  & 0.358 & 0.211 & 0.131 & 0.082 & 0.131 & 0.288 & 0.210   \\
    R2GenCMN~\cite{chen2022cross} & 0.362 & 0.214 & 0.133 & 0.083 &	0.133 &	0.286 & 0.220 \\
    MSAT~\cite{wang2022medical} & 0.362 & \textbf{0.218} & 0.131 & 0.081 & 0.125 & 0.282 & 0.235  \\
    X-RGen (ours) &0.356	&0.216	& \textbf{0.135} & \textbf{0.086} & 0.138 & \textbf{0.294} &0.192  \\
    \midrule
    \multicolumn{8}{c}{generalist models} \\
    \midrule
    R2Gen$^\dagger$ (bs=16) &0.351	&0.199	&0.120	&0.074	&0.132	&0.275	&0.203   \\
    R2Gen$^\dagger$ (bs=96) &0.361	&0.209	&0.126	&0.080	&0.137	&0.281	&0.226   \\
    R2Gen$^\dagger$ (bs=192) & \textbf{0.367} &0.214	&0.133	& \textbf{0.086} & \textbf{0.139} &0.285	&0.238   \\
     X-RGen (ours, bs=16) &0.332	&0.187	&0.113	&0.073	&0.129	&0.263	&0.184   \\
     X-RGen (ours, bs=96) &0.366	&0.211	&0.130	&0.084	&0.137	&0.281	& \textbf{0.257}    \\
     X-RGen (ours, bs=192) & \textbf{0.367}	&0.206	&0.122	&0.076	&0.133	&0.277	&0.215   \\
    \bottomrule
    \end{tabular}%
    }
  \label{tab:Hip}%
\end{table*}%

\begin{table*}[hbpt]
  \centering
  \caption{Comparison with the recent specialised models on Wrist.
  $^\dagger$ means we optimise the model on our merged training dataset while the ``bs'' is the training batch size. All evaluations are conducted on the test set, and a higher value indicates better performance.}
  \resizebox{0.8\linewidth}{!}
  {
    \begin{tabular}{l|ccccccc}
    \toprule
          & BLEU-1 & BLEU-2 & BLEU-3 & BLEU-4 & METEOR & ROUGE-L & CIDEr  \\
    \midrule
    \multicolumn{8}{c}{specialised models} \\
    \midrule
    Transformer~\cite{vaswani2017attention} & 0.339 & 0.203 & 0.133 & 0.086 & 0.120 & 0.301 & 0.129  \\
    R2Gen~\cite{chen2020generating} & 0.359	& 0.214 & 0.139 & 0.093	& 0.135 & 0.299 & 0.288 \\
    R2GenCMN~\cite{chen2022cross} & 0.351 & 0.210 &	0.134 &	0.087 &	0.129 &	0.290 &	0.212  \\
    MSAT~\cite{wang2022medical} & 0.374 & 0.216 & 0.134 & 0.081 & 0.124 & 0.295 & 0.180 \\
    X-RGen (ours) &0.358	&0.214	&0.137	&0.089	&0.142	&0.302	&0.243  \\
    \midrule
    \multicolumn{8}{c}{generalist models} \\
    \midrule
    R2Gen$^\dagger$ (bs=16) &0.351	&0.207	&0.133	&0.085	&0.136	&0.293	&0.217  \\
    R2Gen$^\dagger$ (bs=96) &0.375	&0.215	&0.133	&0.084	&0.144	&0.291	&0.258  \\
    R2Gen$^\dagger$ (bs=192) &0.389	& \textbf{0.238} & \textbf{0.154} & \textbf{0.102} &0.148	& \textbf{0.312} &0.296  \\
     X-RGen (ours, bs=16) &0.342	&0.199	&0.124	&0.079	&0.133	&0.280	&0.229  \\
     X-RGen (ours, bs=96) &0.368	&0.217	&0.138	&0.090	&0.144	&0.298	&0.255   \\
     X-RGen (ours, bs=192) & \textbf{0.390} &0.232	&0.148	&0.097	& \textbf{0.149} &0.299	& \textbf{0.305}  \\
    \bottomrule
    \end{tabular}%
    }
  \label{tab:Wrist}%
\end{table*}%

\begin{table*}[hbpt]
  \centering
  \caption{Comparison with the recent specialised models on Shoulder.
  $^\dagger$ means we optimise the model on our merged training dataset while the ``bs'' is the training batch size. All evaluations are conducted on the test set, and a higher value indicates better performance.}
  \resizebox{0.8\linewidth}{!}
  {
    \begin{tabular}{l|ccccccc}
    \toprule
          & BLEU-1 & BLEU-2 & BLEU-3 & BLEU-4 & METEOR & ROUGE-L & CIDEr  \\
    \midrule
    \multicolumn{8}{c}{specialised models} \\
    \midrule
    Transformer~\cite{vaswani2017attention} & 0.363 & 0.219	& 0.138 & 0.088 & 0.123 & 0.301 & 0.192 \\
    R2Gen~\cite{chen2020generating} & 0.358 & 0.213 & 0.130 & 0.082 & 0.122 & 0.307 & 0.174  \\
    R2GenCMN~\cite{chen2022cross} & 0.348 &	0.210 &	0.129 &	0.082 &	0.119 &	0.297 &	0.134 \\
    MSAT~\cite{wang2022medical} &  0.364 & 0.221 & 0.131 & 0.080 & 0.123 & 0.297 & 0.173 \\
    X-RGen (ours) &0.353	&0.211	&0.133	&0.088	&0.129	&0.304	&0.197 \\
    \midrule
    \multicolumn{8}{c}{generalist models} \\
    \midrule
    R2Gen$^\dagger$ (bs=16) &0.355	&0.212	&0.131	&0.082	&0.132	&0.299	&0.186  \\
    R2Gen$^\dagger$ (bs=96) &0.374	&0.225	&0.142	&0.095	&0.142	&0.297	&0.274  \\
    R2Gen$^\dagger$ (bs=192) &0.380	&0.231	&0.145	& \textbf{0.096} & \textbf{0.144} &0.299	&0.277  \\
     X-RGen (ours, bs=16) &0.350	&0.207	&0.128	&0.084	&0.133	&0.288	&0.220  \\
     X-RGen (ours, bs=96) &0.369	&0.225	&0.145	&0.099	&0.139	& \textbf{0.304} &0.272   \\
     X-RGen (ours, bs=192) & \textbf{0.389}	& \textbf{0.234} & \textbf{0.146} & \textbf{0.096}	&0.141	&0.302	& \textbf{0.287} \\
    \bottomrule
    \end{tabular}%
    }
  \label{tab:Shoulder}%
\end{table*}%

\end{document}